%% file: acl_latex.tex
\pdfoutput=1

\documentclass[11pt]{article}

\usepackage[preprint]{acl}

\usepackage{balance}
\usepackage{times}
\usepackage{latexsym}
\usepackage{ulem}
\usepackage{amssymb}
\usepackage[T1]{fontenc}

\usepackage[utf8]{inputenc}

\usepackage{algorithm2e}

\usepackage{microtype}

\usepackage{inconsolata}

\usepackage{graphicx}

\usepackage{booktabs} 
\usepackage{multirow}
\usepackage{colortbl}
\usepackage{xspace}
\usepackage{tabularx}
\usepackage{longtable}
\usepackage{amsmath}
\usepackage{rotating}
\usepackage{subcaption}

\usepackage[capitalize]{cleveref}
\crefname{section}{Sec.}{Secs.}
\Crefname{section}{Section}{Sections}
\Crefname{table}{Table}{Tables}
\crefname{table}{Tab.}{Tabs.}
\definecolor{myred}{rgb}{0.752, 0.004, 0.004}
\usepackage{enumitem}
\setenumerate{itemsep=0pt,partopsep=0pt,parsep=0pt,topsep=2pt}

\newcommand{\pq}{probe-Query\xspace}
\newcommand{\name}{ActQKV\xspace}

\newcommand{\basellama}{LLaMA3-8B-inst\xspace}

\title{Activation-aware Probe-Query: \\Effective Key-Value Retrieval for Long-Context LLMs Inference}

\author{
\textbf{Qingfa Xiao$^{1}$, Jiachuan Wang$^{2}$, Haoyang Li$^{3}$, Cheng Deng$^{1}$, Jiaqi Tang$^{1}$} \\ 
\textbf{Shuangyin Li$^{4}$, Yongqi Zhang$^{1}$, Jun Wang$^{5}$, Lei Chen$^{1,2}$} \\\\
$^1$ The Hong Kong University of Science and Technology (Guangzhou) \\ $^2$ The Hong Kong University of Science and Technology \\ $^3$ The Hong Kong Polytechnic University \\ $^4$ South China Normal University \\$^5$ University College London \\\\
}

\begin{document}
\maketitle

\input{latex/tex/abstract}

\input{latex/tex/introduction}

\input{latex/tex/related_works}

\input{latex/tex/background}

\input{latex/tex/methods}

\input{latex/tex/experiments}

\input{latex/tex/conclusion}

\input{latex/tex/limitations}

\section*{Ethics Statement}
Throughout the development and execution of this work, we strictly adhered to ethical guidelines established by the broader academic and open-source community. All datasets and models utilized are publicly available. There are no conflicts of interest among the authors involved in this research. Our approach aligns with ethical AI practices, prioritizing trust, accountability, and responsible research.

\bibliography{custom}

\input{latex/tex/appendix}

\end{document}

%% file: latex/tex/abstract.tex
\begin{abstract}

Recent advances in large language models (LLMs) have showcased exceptional performance in long-context tasks, while facing significant inference efficiency challenges with limited GPU memory. Existing solutions first proposed the sliding-window approach to accumulate a set of historical \textbf{key-value} (KV) pairs for reuse, then further improvements selectively retain its subsets at each step.
However, due to the sparse attention distribution across a long context, it is hard to identify and recall relevant KV pairs, as the attention is distracted by massive candidate pairs. Additionally, we found it promising to select representative tokens as \pq in each sliding window to effectively represent the entire context, which is an approach overlooked by existing methods.
Thus, we propose \textbf{\name}, a training-free, \textbf{Act}ivation-aware approach that dynamically determines probe-\textbf{Q}uery and leverages it to retrieve the relevant \textbf{KV} pairs for inference.
Specifically, \name monitors a token-level indicator, Activation Bias, within each context window, enabling the proper construction of \pq for retrieval at pre-filling stage. 
To accurately recall the relevant KV pairs and minimize the irrelevant ones, we design a dynamic KV cut-off mechanism guided by information density across layers at the decoding stage. Experiments on the Long-Bench and $\infty$ Benchmarks demonstrate its state-of-the-art performance with competitive inference quality and resource efficiency. 
\end{abstract}

%% file: latex/tex/introduction.tex
\section{Introduction}

\begin{figure}[!ht] 
    \centering
        \includegraphics[width=0.45\textwidth]{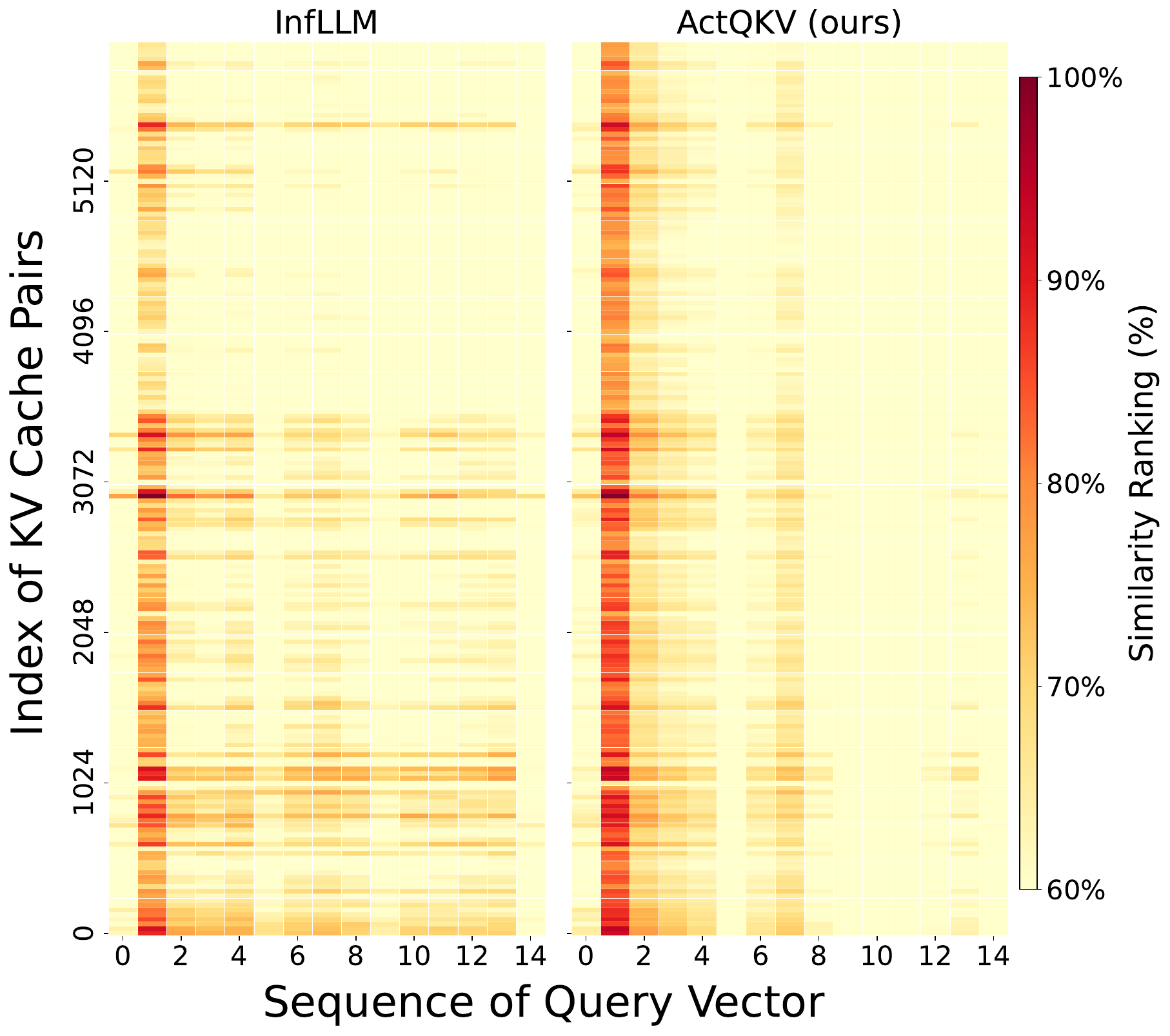}
        \includegraphics[width=0.45\textwidth]{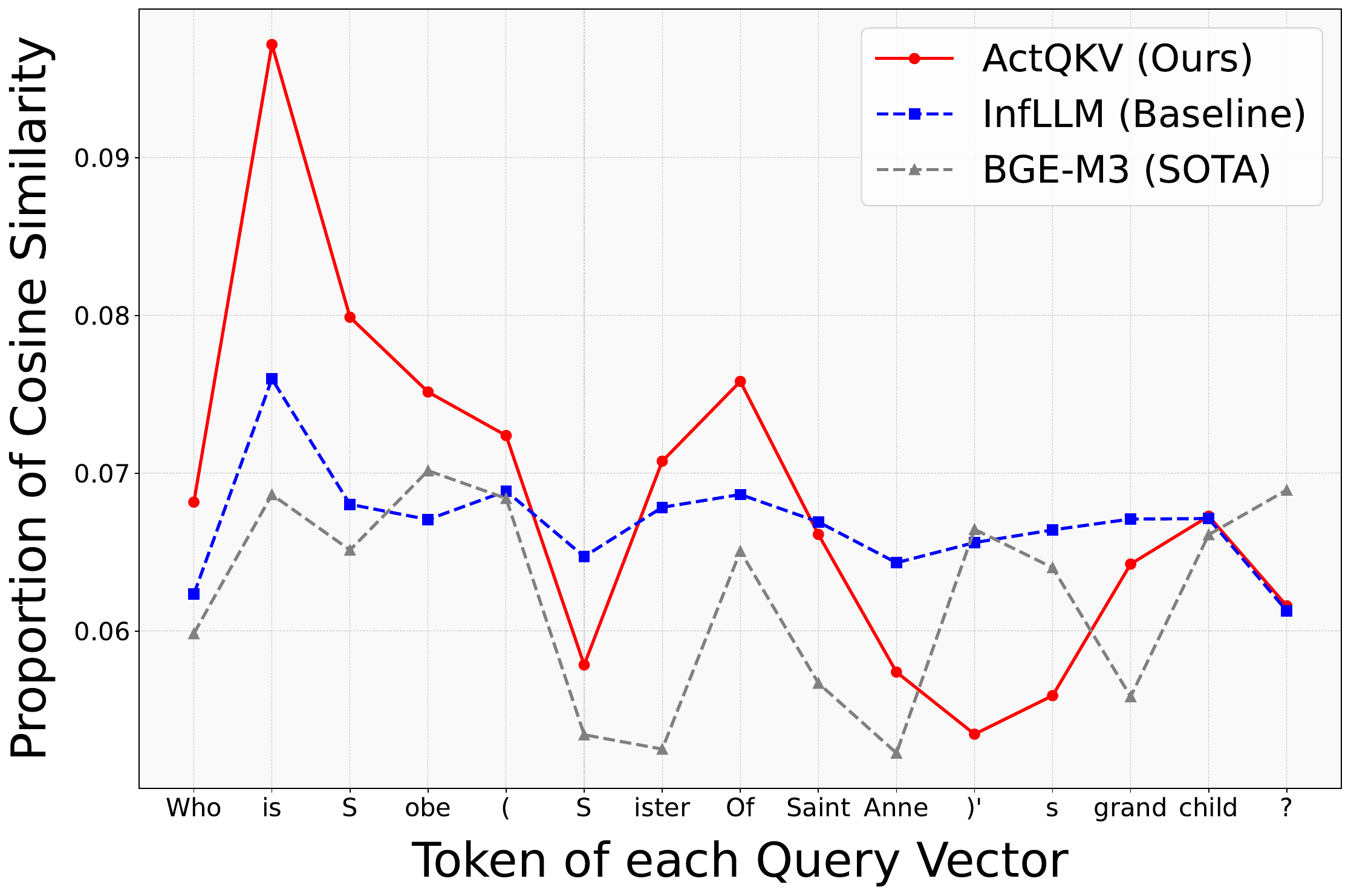}
    \vspace{-0.2em}
    \caption{Visualization of query vector status within \pq compared between \name and InfLLM: "Who is Sobe (Sister of Saint Anne)’s Grandchild?". We simply display the states of 15 tokens from a window of size 256 in the last transformer layer. The \pq generated by our \name aligns more closely with the SOTA embedding model BGE-M3~\cite{chen-etal-2024-m3}. In contrast, InfLLM generates evenly distributed similarities across the context, neglecting the prioritization of anchor tokens compared to our approach.
    } 
    \vspace{-1em}
    \label{fig:motivation}
\end{figure}

With the emergence of large language models (LLMs) capable of handling extended context lengths~\cite{ wang2024beyond,achiam2023gpt, dubey2024llama}, researchers are leveraging their advanced information understanding and filtering abilities to tackle various downstream tasks, including web-based search chatbot~\cite{Semnani2023WikiChat} and document-level question answering (QA)~\cite{lewis2020retrieval}. 
Inevitably, the context length has increased significantly, even surpassing the models’ context limitations. 
However, the computational complexity of attention mechanism~\cite{vaswani2017attention} grows quadratically $O(N^2)$ with the context length $N$ during inference. Specifically, each token from context will be embedded into Query (Q) and interactive with Key (K) and Value (V) embedded from all the $N$ tokens using attention weights, making the whole time and memory complexity $O(N^2)$ for the process.
Even worse, during inference, new tokens are generated one by one while each generation triggers a $O(N^2)$ computation, leading to an $O(N^2+MN^2)$ to generate an output of length $M$.
Therefore, efficiency is a critical challenge in the deployment of long-context LLMs~\cite{li2024survey}.

To handle this issue, the sliding window mechanism has been proposed to segment the input sequence into content blocks and incrementally convert them into a  key-value (KV) cache for reuse~\cite{beltagy2020longformer}. 
During inference, the model computes the KV vectors only for the current window and integrates them with the existing KV cache, thereby reducing redundant KV computations, leading to an $O(N^2+MN)$ complexity.
Building on this mechanism, recent works~\cite{infllm,liu2024retrievalattention} focus on retrieving top-$k$ relevant KV pairs in conjunction with current tokens for preserving long-term contextual dependencies, where further reduces the complexity to $O(kN+kM)$.
In this process, the queries from current window are typically compressed as a \textbf{\pq} for relevant KV retrieval. However, this \pq setting often fails to highlight those anchor tokens with critical activation signals, which are rare and essential to represent long context within the sliding window.

To address this challenge, we first investigate the similarity relationship between the composition of the \pq and KV cache.
Under sparse attention patterns (see upper of Fig.~\ref{fig:motivation}), the query vectors generated by InfLLM (the left) are disordered. In this scenario, each query vector influences the semantics of \pq, which makes the combined representation nondescript. To clearly demonstrate this nondescript (see bottom of Fig.~\ref{fig:motivation}), the {blue line}  employs a widely used mean pooling technology along KV dimension to represent the \pq. It is evident that the \pq fails to capture the distinctions because attention is distracted by all tokens instead of focusing on the anchors.
Therefore, such a nondescript \pq is hard to represent semantic of question and unsuitable for effective KV retrieval.

Motivated by these observations, we argue that only a subset of anchor tokens within the context window plays a dominant role in representing \pq for retrieval. In this paper, we propose \name, a training-free method that incorporates sliding window attention, which mainly involves two stages: matching and recall of relevant KV pairs. \textbf{In KV matching stage}, we construct the \pq for each context window to retrieve the relevant KV pairs in a streaming manner. To effectively estimate the anchor tokens during inference, we employ a window-level activation-aware strategy to monitor the fluctuation of query values for each token. Recognizing that the scarce outlier features is a critical factor affecting model performance~\cite{wang2024learning,wu2024retrieval}, we designate activated query vectors with prominent activation bias to dominate the representation of \pq for accurate retrieval, as shown in {red line} of Figure~\ref{fig:motivation}. 
\textbf{In KV recall stage}, due to the irregular distribution of KV pairs across layers, a fixed threshold often fails to yield optimal retrieval results. In particular, the decoding stage, which is highly sensitive to factual correctness, can be adversely affected by irrelevant KV pairs, potentially leading to hallucinations and degrading the overall quality of the generated text. Therefore, we introduce a KV cut-off mechanism that dynamically adjusts the number of selected pairs based on information density of each layer. Under a constrained KV budget, this mechanism enhances the recall of relevant KV pairs while reduces the introduction of irrelevant ones.

Our contributions are summarized as follows:
\vspace{-0.4em}
\begin{itemize}
  \item Motivated by attention distraction phenomenon, we introduce an activation-aware \pq that efficiently emphasizes anchor tokens essential for accurately matching KV pairs. It is the first exploration to extract long-context representations for KV retrieval without training.
  \vspace{-0.4em}
  \item To further eliminate irrelevant KV pairs and recall the relevant, we design a dynamic KV cut-off mechanism guided by information density across layers during the decoding stage. This method effectively enhances the model's factual filtering ability for reasoning QA.
  \vspace{-0.4em}
  \item Our \name outperforms existing SOTA KV retrieval-based methods with just 2K KV budget on two benchmarks, achieving up to a 16× KV reduction and 10.4\% accuracy improvement compared to using the full cache setting with a 2K budget on LongBench.
\end{itemize}

%% file: latex/tex/related_works.tex
\section{Related Works}

\textbf{KV cache retrieval}~\cite{adnan2024keyformer,zhang2023h2o,xiao2025duoattention} has become a critical optimization strategy aimed at reducing memory usage, minimizing inference latency and improving overall throughput in long-context LLMs inference. 

Recent studies employ a sliding window mechanism to address challenges in long-text inference, where tokens outside the window are stored in the cache and only used when needed for the current window. 
To accelerate the retrieval of essential KV, several approaches have proposed index-based methods that organize and access the KV cache at the block or cluster level, enabling efficient querying and extraction. InfLLM~\cite{infllm} maintains the full KV cache in blocks and uses a hierarchical storage strategy to facilitate long-sequence processing. This framework employs CPU-GPU memory orchestration, keeping essential KV and computational units in GPU memory while offloading less frequently accessed units to CPU memory. Q-LLM~\cite{qllm} enhances long-sequence processing by prioritizing memory related to task descriptions. This approach mimics human reading behavior: first reading the question, then searching for the answer in the context. 

In contrast to methods which use uniform KV block sizes, TokenSelect\cite{tokenselect} is based on the observation of sparsity in non-continuous attention patterns. It uses the Query-Key dot product to assess the importance of each KV cache stored at the token level. For each query, they dynamically calculates the importance of past KV caches per head at the token level and selects the most important tokens through a soft voting mechanism across heads. EM-LLM~\cite{emllm} dynamically segments incoming tokens into episodic events, employing a hybrid retrieval mechanism that combines semantic similarity matching with temporal context to efficiently access relevant KV cache segments. Additionally, some researchers focus on KV cache budget allocation across layers~\cite{cai2024pyramidkv, yang-etal-2024-pyramidinfer} and heads~\cite{feng2024ada, fu2025not} due to the hierarchical architecture of LLMs.

Most methods overlook the importance of probes for retrieval, especially given the fact that LLMs are not optimized for retrieval tasks. Therefore, this realization inspires our further exploration of \pq construction in this paper.

%% file: latex/tex/background.tex
\section{Background}
In this section, we first introduce the two stages of inference for long-context LLMs using sliding window attention with KV cache (in \cref{sec:window}), and then define the problem of KV Retrieval (in \cref{sec:problem}).

\subsection{Sliding Window Attention with KV Cache}
\label{sec:window}
Given an input sequence $\mathbf{X}$, the generation of the output sequence $\mathbf{Y}$ during LLMs inference can be divided into two stages: pre-filling the input $\mathbf{X}$ and decoding the output $\mathbf{Y}$.

\begin{figure*}[!ht] 
    \centering
        \includegraphics[width=1\textwidth]{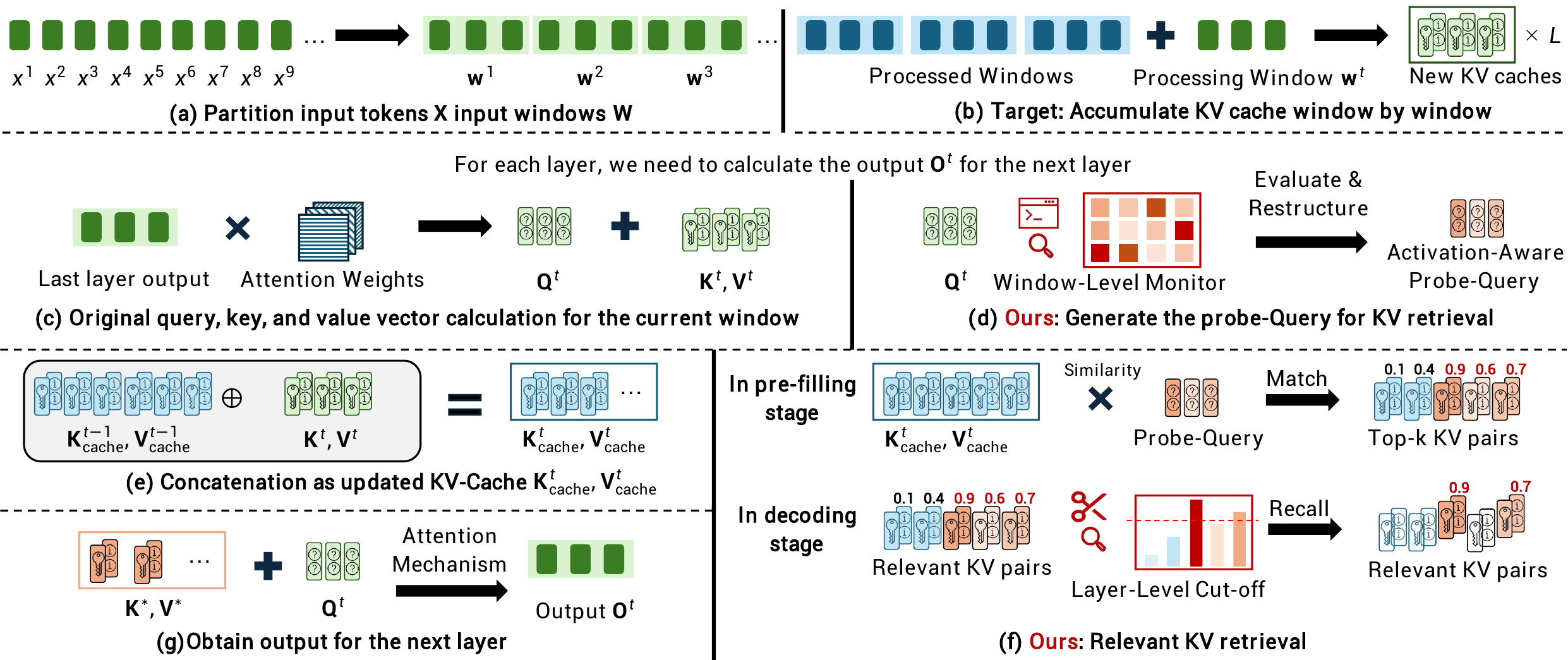}
    \caption{\text{Illustration of our \name}. Sliding window attention stores historical KV pairs in a cache and reuses them for subsequent window inference. Based on this, \name first identifies the anchor tokens within the window and then constructs the activation-aware \pq. This \pq is subsequently used to retrieve the top-k relevant KV pairs from the cache during the pre-filling stage. During the decoding stage, the cut-off mechanism dynamically adjusts the number of recalled KV pairs based on the distribution of key-values at each layer, ensuring the inclusion of relevant pairs while minimizing the influence of irrelevant ones. The cache can be stored in the CPU and transferred to the GPU when needed. All our contributions are highlighted in {red}. }
    \label{fig:framework}
\vspace{-0.5em}
\end{figure*}

To handle long sequences input of tasks, exiting works~\cite{stream-llm, infllm, qllm} use sliding window attention to process the text iteratively. In this mechanism, the lengthy input sequence $\mathbf{X}$ is partitioned into $T$ windows, denoted as $\mathbf{W} = \{ \mathbf{w}^{1}, \dots, \mathbf{w}^{T}\}, \mathbf{W}\in \mathbb{R}^{T \times m}$ and $m$ indicates the window size (see~\cref{fig:framework}(a)). To reduce computational costs, the model processes each window sequentially  and stores the historical key-value pairs in a cache (i.e., $\mathbf{K}_{\text{cache}}$ and $\mathbf{V}_{\text{cache}}$) for future reuse (see~\cref{fig:framework}(b)).

\textbf{During $t$-th pre-filling step  ($t \leq T$)}, the model utilizes the KV cache $\mathbf{K}^{t-1}_{\text{cache}}$ and $\mathbf{V}^{t-1}_{\text{cache}}$ from the historical sequence $\mathbf{W}[:t-1]$ to compute the attention output $\mathbf{O}^{t} \in \mathbb{R}^{m \times d}$ for the current $m$ window tokens $\mathbf{w}^{t} \in \mathbb{R}^{m}$ as follows:
\begin{equation}
\mathbf{O}^{t} = \text{Attention}\left( \mathbf{Q}^{t}, \left[ \mathbf{K}^{t}, \mathbf{K}^{t-1}_{\text{cache}} \right], \left[ \mathbf{V}^{t}, \mathbf{V}^{t-1}_{\text{cache}} \right] \right),
\label{eq:prefilling}
\end{equation}
where the triplet $\mathbf{Q}^{t}=\{\mathbf{q}^{t}_i\}^m_{i=1}$, $\mathbf{K}^{t}=\{\mathbf{k}^{t}_i\}^m_{i=1}$, $\mathbf{V}^{t}=\{\mathbf{v}^{t}_i\}^m_{i=1} \in \mathbb{R}^{m \times d}$ represents the generated attention vectors, each corresponds to $m$ tokens with $d$ hidden dimensions. To further save GPU memory, current methods select partial KV cache $\mathbf{K}^{*}$ and $\mathbf{V}^{*}$ for inference, denoted as:
\begin{equation}
\mathbf{O}^{t} = \text{Attention}\left( \mathbf{Q}^{t}, \left[ \mathbf{K}^{t}, \mathbf{K}^{*} \right], \left[ \mathbf{V}^{t}, \mathbf{V}^{*} \right] \right),
\label{eq:prefilling}
\end{equation}
where $\mathbf{K}^{*} \subseteq \mathbf{K}^{t-1}_{\text{cache}}$ and $\mathbf{V}^{*} \subseteq \mathbf{V}^{t-1}_{\text{cache}}$. 

\textbf{During $t$-th decoding step  ($t > T$)}, the model generates the output sequence $\mathbf{Y}$ token-by-token. Unlike pre-filling, the model uses only one single query vector $\mathbf{q}^{t} \in \mathbb{R}^{1 \times d}$ along with corresponding key and value vectors $\mathbf{k}^{t}, \mathbf{v}^{t} \in \mathbb{R}^{1 \times d}$ to predict one next token $y^{t} \in \mathbf{Y}$ in each step. Its corresponding attention output $\mathbf{o}^{t} \in \mathbb{R}^{1 \times d}$ can be computed as:
\begin{equation}
\vspace{-3pt}
\begin{aligned}
\mathbf{o}^{t} = \text{Attention}\left( \mathbf{q}^{t}, \left[ \mathbf{k}^{t}, \mathbf{K}^{*} \right], \left[ \mathbf{v}^{t}, \mathbf{V}^{*} \right] \right).
\end{aligned}
\end{equation}

\textbf{After the $t$-th step}, the newly generated key-value pairs will be stored in the cache (see~\cref{fig:framework}(e)), updating it as demonstrated below:
\begin{equation}
\begin{aligned}
\mathbf{K}^{t}_{\text{cache}},\mathbf{V}^{t}_{\text{cache}} = \mathbf{K}^{t-1}_{\text{cache}}\cup \mathbf{K}^{t}, \mathbf{V}^{t-1}_{\text{cache}}\cup \mathbf{V}^{t},
\end{aligned}
\end{equation}
where $\cup$ denotes the concatenation operation and the tensors of cache can be saved in either CPU or GPU memory. In general, saving in the CPU can significantly reduce the memory usage of the GPU. Note that $\mathbf{K}^t=\mathbf{k}^t$ and $\mathbf{V}^t=\mathbf{v}^t$ are $1\times d$ dimensions during decoding.

\subsection{Problem Setting}
\label{sec:problem}
During long-context inference in LLMs, the historical key-value pairs are essential for maintaining long-range dependencies and overcoming window size limitations. Given a cache comprising $\mathbf{K}^{t-1}_{\text{cache}}$ and $\mathbf{V}^{t-1}_{\text{cache}}$, the objective of KV retrieval is to identify the top-$k$ relevant subset $\mathbf{K}^{*}$ and $\mathbf{V}^{*}$ using the \pq $\mathbf{Q}^{t}_{\text{probe}}$ for the $t$-th inference step~\cite{infllm,emllm,tokenselect}, as described below:
\begin{equation}
\begin{aligned}
&\mathbf{K}^{*},\mathbf{V}^{*} =\mathbf{K}^{t-1}_{\text{cache}}[I^*],\mathbf{V}^{t-1}_{\text{cache}}[I^*], \\
I^* = \arg&\max_{I \subset [m],\atop |I| = k} \sum_{i \in I} \left( \frac{\mathbf{Q}^{t}_{\text{probe}} \cdot {\mathbf{K}^{t-1}_{\text{cache}}}[i]^\top}{\|\mathbf{Q}^{t}_{\text{probe}}\| \times \| \mathbf{K}^{t-1}_{\text{cache}}[i] \|} \right),\\ 
&\quad\quad\quad\quad\quad\quad\quad[m] = \{1,2,\ldots, m\},&
\end{aligned}
\end{equation}
where $\mathbf{Q}^{t}_{\text{probe}} \in \mathbb{R}^{1 \times d}$ denotes the overall representation of window context $\mathbf{w}^{t}$ and $k$ is the number of selected KV. These two factors significantly impact the factual relevance of the retrieved KV index $I^*$ for each transformer layer inference.

%% file: latex/tex/methods.tex

\section{Methods}

In this section, we first present the overall framework of our \name, as illustrated in \cref{fig:framework}. We then demonstrate our two-stage approach: the Activation-aware Probe-Query Construction for KV matching (in \cref{sec:pq})and the Dynamic KV Cut-off Mechanism for KV recall (in \cref{sec:co}).

\subsection{Activation-Aware Probe-Query}
\label{sec:pq}
To identify the relevant KV pairs, we leverage the query vectors of each window to construct the attention-aware \pq for retrieval.
The primary distinction between our activation-aware \pq and other representation methods lies in the emphasis on identifying anchor tokens that effectively represent the entire context of the window for KV matching. The main challenge is to accurately distinguish and activate these tokens.

Formally, given a subset of context $\mathbf{w}^t = \{x^{t}_{1}, \dots, x^{t}_{m}\}$ extracted from a long sequence $\mathbf{W}$, we obtain the hidden states $\{z^{t}_{i}\}_{i=1}^m=\{f(x^{t}_{i})\}_{i=1}^m$ at each transformer layer, where $m$ denotes the window size and $f$ denotes the function mapping tokens to corresponding states. Intuitively, hidden states that deviate significantly from their statistical mean (i.e., $\mathbf{\bar{z}}^t$) can be considered that they are from anchor tokens compared to others. Specifically, token ${x}^{t}_{1}$ is deemed more essential than ${x}^{t}_{2}$ for the quality of generation, as indicated by previous works~\cite{wang2024learning, sun2024massive,pang-etal-2024-anchor}, if:
\begin{equation}
\begin{aligned}
\| \mathbf{\bar{z}}^t - f(x^{t}_{1}) \| > \| \mathbf{\bar{z}}^t - f(x^{t}_{2}) \|,
\label{eq:activation}
\end{aligned}
\end{equation}
where $||\cdot||$ is distance metrics.

Building on the aforementioned paradigm Eq.~\ref{eq:activation}, we propose an \textbf{Activation Bias} to distinguish the importance of each query vector within a window context. For the query vectors of the $t$-th pre-filling window $\mathbf{Q}^{t} = \{\mathbf{q}^{t}_{1}, \dots, \mathbf{q}^{t}_{m}\}$ in each layer, we first compute the token-level bias $\mathbf{\Phi}^t = \{\mathbf{\phi}^{t}_{1}, \dots, \mathbf{\phi}^{t}_{m}\}$, with $\mathbf{\Phi}^t \in \mathbb{R}^{m \times d}$, to estimate the energetic degree within $\mathbf{Q}^{t}$ as follows:

\begin{equation}
\begin{aligned}
\mathbf{\phi}^t_j = \frac{(\mathbf{q}^{t}_{j} - {\mathbf{\bar{z}}^t})^2}{\mathbf{\sigma}^2}&,
\end{aligned}
\end{equation}
where $\mathbf{\sigma}^2$ and $\mathbf{\bar{z}}^t \in \mathbb{R}^{1 \times d}$ represent the variance and mean of the query vectors respectively, computed as follows:
\begin{equation}
\begin{aligned}
\mathbf{\sigma}^2 = \frac{\sum_{i=1}^{t} \sum_{j=1}^{m} \left( \mathbf{q}^i_j - \mathbf{\bar{z}}^t \right)^2}{mt-1},\mathbf{\bar{z}}^t =&\frac{\sum_{i=1}^{t}\sum_{j=1}^{m} \mathbf{q}^i_j}{mt}.
\end{aligned}
\end{equation}

Based on the above estimated degree, we can construct the \pq $\mathbf{Q}^{t}_{\text{probe}}$ for KV matching by reassigning the activated weights of each query vector according to the activation bias $\mathbf{\Phi}^t$:
\begin{equation}
\begin{aligned}
\mathbf{Q}^{t}_{\text{probe}} = \sum_{j=1}^{m}\frac{\Vert \mathbf{\phi}^t_j \Vert_1}{\Vert \mathbf{\Phi}^t \Vert_1}\mathbf{q}^{t}_{j}.
\end{aligned}
\end{equation}
Our object is to enhance the weight of query vectors for those anchor tokens. With this activated \pq, we can match more precise KV pairs $\mathbf{K}^{*}$ and $\mathbf{V}^{*}$ that contain semantically relevant information for pre-filling stage Eq.~\ref{eq:prefilling}.

\subsection{Dynamic KV Cut-off Mechanism}
\label{sec:co}
During the decoding stage, the quality of the predicted answer greatly depends on the top-$k$ relevant pairs $\mathbf{K}^{*}$ and $\mathbf{V}^{*}$. However, due to the sparse and irregular attention pattern across each layer, the selection of $k$ KV pairs is highly sensitive to the \pq $\mathbf{Q}^{t}_{\text{probe}}=\mathbf{q}^{t}$. Therefore, we propose a KV cut-off mechanism to dynamically determine $k$ based on information density assessment for $L$ transformer layers. Compared to the preset threshold, this mechanism dynamically removes redundant KV pairs and improves the recall of relevant ones within a limited KV budget.

In the $t$-th decoding step, we first calculate the similarity scores $\mathbf{S}^\ell = \{s_{1}^\ell, \dots, s_{n}^\ell \}$ between the \pq $\mathbf{Q}^{t}_{\text{probe}}$ and the cache of key vectors $\mathbf{K}^{t-1}_{\text{cache}}$ for the $\ell$-th transformer layer, where $n = |\mathbf{K}^{t-1}_{\text{cache}}|$. The similarity scores are computed using cosine similarity as follows:
\begin{equation}
s_i^\ell = \frac{\mathbf{Q}^{t}_{\text{probe}} \cdot \mathbf{K}^{t-1}_{\text{cache}}[i]}{\|\mathbf{Q}^{t}_{\text{probe}}\| \times \|\mathbf{K}^{t-1}_{\text{cache}}[i]\|}.
\label{eq:cosine_similarity}
\end{equation}
Then, we apply the softmax function to normalize them and convert them into probabilities.

Based on the similarity distribution $\mathbf{S}^\ell$, we define the information density $\Theta^\ell$ for the $\ell$-th layer using the entropy function as follows:
\begin{equation}
\begin{aligned}
\Theta^\ell= - \sum_{i=1}^{n} & \frac{e^{s_i^\ell}}{\sum_{j=1}^{n} e^{s_j^\ell}} \log \left( \frac{e^{s_i^\ell}}{\sum_{j=1}^{n} e^{s_j^\ell}} \right),
\label{eq:entropy}
\end{aligned}
\end{equation}
where a uniform distribution results in a higher information density $\Theta^\ell$ compared to more concentrated distributions.


Now with the information density, we focus on dynamically assigning the budget instead of  a fixed value  $k$  for each layer. Given a total budget $\text{B}_{kv}$, we process from shallow to deep layers in the order of transformer computation to avoid decoding delays. Consequently, for the $\ell$-th layer in the  $t$-th decoding step, the budget $\text{B}^\ell$ can be estimated as follows:
\begin{equation}
\begin{aligned}
\text{B}^\ell = \frac{{\Theta}^\ell}{{{\Theta}^\ell}+{\sum_{j=\ell+1 }^{L}{\bar{\Theta}^j}}}\times \text{B}_{kv},
\label{eq:budget}
\end{aligned}
\end{equation}
where $\text{B}_{kv}$ is initialized as $L \times k$ and updated by $\text{B}_{kv} \leftarrow \text{B}_{kv}- \text{B}^\ell$ after  processing 
 the $\ell$-th layer, and $\bar{\Theta}^j$ denotes the mean $\Theta^\ell$ for the remaining unprocessed layers. In this part, we aim to assign a larger budget to layers with higher information density, where many KV pairs are potentially relevant to the \pq $\mathbf{Q}^{t}_{\text{probe}}$ for the $t$-th decoding step. Conversely, for layers with lower density, the relevant KV pairs with higher similarity are more prominent, making 
 the irrelevant pairs  more likely to be discarded. Based on the above Eq.~\ref{eq:budget}, the denominator, which adds  $\Theta^\ell$  to the cumulative average density  $\sum_{j=\ell+1}^{L} \bar{\Theta}^j$  of the remaining layers, quantifies the overall contribution of both the current and subsequent layers. A higher ratio indicates that the current layer holds a more significant portion of the relevant KV pairs, justifying a larger allocation. Compared to using a fixed threshold for retrieval, this dynamic KV cut-off mechanism eliminates redundant KV pairs and improves the recall of relevant ones within the limited KV budget.

In summary, we present our two-stage method separately, where the activation-aware \pq module guarantees the quality of historical KV pairs and the cut-off mechanism effectively utilizes them. The entire process is depicted in Algorithm~\ref{alg} as shown below:

\RestyleAlgo{ruled}
\begin{algorithm}[!h]
\small
\SetAlgoLined
\SetKwInOut{Input}{Input}
\SetKwInOut{Output}{Output}
\caption{Effective KV Retrieval for Long-context LLMs Inference}\label{alg}
\Input{
    \( L \): Total number of transformer layers;
    \( \mathbf{Q}^{t}_{\text{probe}} \): Probe-Query for the \( t \)-th step;\quad 
    \( \mathbf{K}^{t-1}_{\text{cache}} \): Cache of key vectors for the $t-1$-th step;
    \( \text{B}_{kv} \): Initial KV  budget (\( L \times k \))
}
\Output{
    \( \mathbf{K}^{*} \) and \( \mathbf{V}^{*} \): Selected KV pairs for inference
}

\LinesNumbered

\For{\( \ell \gets 1 \) \KwTo \( L \)}{
    \For{\( i \gets 1 \) \KwTo \( n \)}{
        \( s_i^\ell \gets \frac{\mathbf{Q}^{t}_{\text{probe}} \cdot \mathbf{K}^{t-1}_{\text{cache}}[i]}{\|\mathbf{Q}^{t}_{\text{probe}}\| \times \|\mathbf{K}^{t-1}_{\text{cache}}[i]\|} \);
    }
    \( \mathbf{P}^\ell \gets \text{Softmax}(\mathbf{S}^\ell) \);
    \( \Theta^\ell \gets -\sum_{i=1}^{n} P(s_i^\ell) \log P(s_i^\ell) \);
}

\For{\( \ell \gets 1 \) \KwTo \( L \)}{
    \( \text{B}^\ell \gets \frac{\Theta^\ell}{\Theta^\ell + \sum_{j=\ell+1}^{L} \bar{\Theta}^j} \times \text{B}_{kv} \);
    \( \text{B}_{kv} \gets \text{B}_{kv} - \text{B}^\ell \);
}

\For{\( \ell \gets 1 \) \KwTo \( L \)}{
    Sort \( \mathbf{K}^{t-1}_{\text{cache}} \) based on \( \mathbf{P}^\ell \) in descending order;
    Select top \( \text{E}^\ell \) KV pairs;
    Add selected KV pairs to \( \mathbf{K}^{*} \) and \( \mathbf{V}^{*} \);
}

\Return \( \mathbf{K}^{*} \), \( \mathbf{V}^{*} \);

\end{algorithm}

%% file: latex/tex/experiments.tex
\input{latex/tables/long}
\vspace{-0.5em}
\section{Experiments}
In this section, we first present the experimental setup of this paper (in \cref{sec:setup}). Then we demonstrate the logical reasoning and factual retrieval ability of our \name in long-context inference through two widely-used benchmark (in \cref{sec:results}). Finally, we conduct the ablation study (in \cref{sec:ablation}) and reveal the influence of our method (in \cref{sec:analysis}).

\subsection{Experimental Setup}
\label{sec:setup}

\paragraph{Datasets and Implementation Details.}
We utilize 21 tasks from two widely used long document benchmarks: Long-Bench~\cite{longbench} and $\infty$-Bench~\cite{infinitebench} for evaluation. Specifically, 
Long-Bench has a 95\% sequence length of 32K, while $\infty$-Bench averages about 122K in sequence length.  We utilize \basellama~\cite{llama3} and  Qwen2.5-7B-Instruct~\cite{qwen2.5} as our base models with maximum input lengths of 8K and 32K, respectively.
In each inference step, we reuse only 2K KV pairs and store the remaining pairs in the Cache Management system, following the settings of InfLLM. This approach consumes approximately 19 GB of VRAM in our experiments. Inspired by previous works, we retain 64 attention sinks and 512 KV pairs from current context, and adapt the task description into \pq. Consequently, the budget for retrieved KV $k$ is 1,472. These KV pairs are organized into 46 chunks, with each chunk containing 32 pairs. The sliding window size is set to 256. {More details about the datasets and experimental setup is available in ~\cref{sec:alg}.}

\paragraph{Baseline Methods} The objective of \name is to effectively retrieve key-value pairs for long-context inference in LLMs. To achieve this, we evaluate two prominent baseline methods: \textbf{(a)} static KV selection and \textbf{(b)} KV retrieval. \textbf{(a):} Infinite~\cite{infinite-llm} employs global and local attention masks to broaden the attention scope, while Stream~\cite{stream-llm} ensures efficient inference by retaining attention sinks and KV pairs from recent tokens. \textbf{(b):} InfLLM~\cite{stream-llm} searches for KV pairs associated with the currently processed tokens, enabling the capture of long-distance dependency relationships. QLLM~\cite{qllm} focuses on KV memory relevant to the task description to process long sequences.  TokenSelect (TSLLM)~\cite{tokenselect} incorporates the token-level weight of  KV cache per-head for KV retrieval. EMLLM~\cite{emllm} integrates key aspects of human episodic memory and event cognition into KV cache. Notably, all the methods described above are \textbf{training-free}.

\subsection{Main Experiment Results}
\label{sec:results}

We first utilize Long-Bench to evaluate the long-context reasoning capabilities of \name, and then test the fact retrieval ability using $\infty$-Bench.
{We report the results based on Llama-3-8B-Instruct, and the others can be found in \cref{sec:experiments}.}

\paragraph{Long-Bench.} We present the results in \cref{tab:llama3}. (1) \name achieves an average score of 49.40, surpassing the full context setting (31K tokens) by 4.67 points while utilizing only 2K tokens. This highlights the \textbf{efficiency} of its key-value retrieval method in handling long-context inference with a significantly smaller KV budget. (2) Compared to the static KV selection methods Infinite and Stream, \name excels in capturing critical information required for reasoning tasks. (3) In comparison to SOTA KV retrieval methods such as TSLLM and EMLLM, our activation-aware retrieval approach achieves the best results, with improvements of +5.8\% and +4.6\%, respectively. Notably, for tasks like 2WikiMQA and Musique, \name shows substantial gains, demonstrating the effectiveness of activation-aware retrieval in capturing long-term dependencies by recalling fewer KV pairs (e.g., only with 80\% and 50\% budget).

\input{latex/tables/infty}
 \vspace{-0.5em}
\paragraph{$\infty$-Bench.} Each sample in this benchmark has almost infinite length (avg. 122K), where the key lies in whether factual evidence can be found from the context. As shown in \cref{tab:llama3-infinitybench}, our \name obtains the best result 58.43 and outperforms the SOTA KV retrieval methods even with a smaller KV budget. Especially compared to the token-level retrieval method TSLLM, our approach sets the minimum retrieval unit as a chunk. Although larger chunks may seem less granular, our \pq effectively compensates for this, enhancing 3.5\% performance while simultaneously reducing both time and space complexity from $O(N)$ to $O(m)$. This demonstrates that our method can efficiently recall relevant KV pairs even with coarser granularity.

\input{latex/tables/ablation_study}

\begin{figure*}[!ht] 
\vspace{-0.5em}
    \centering
        \includegraphics[width=1.0\textwidth]{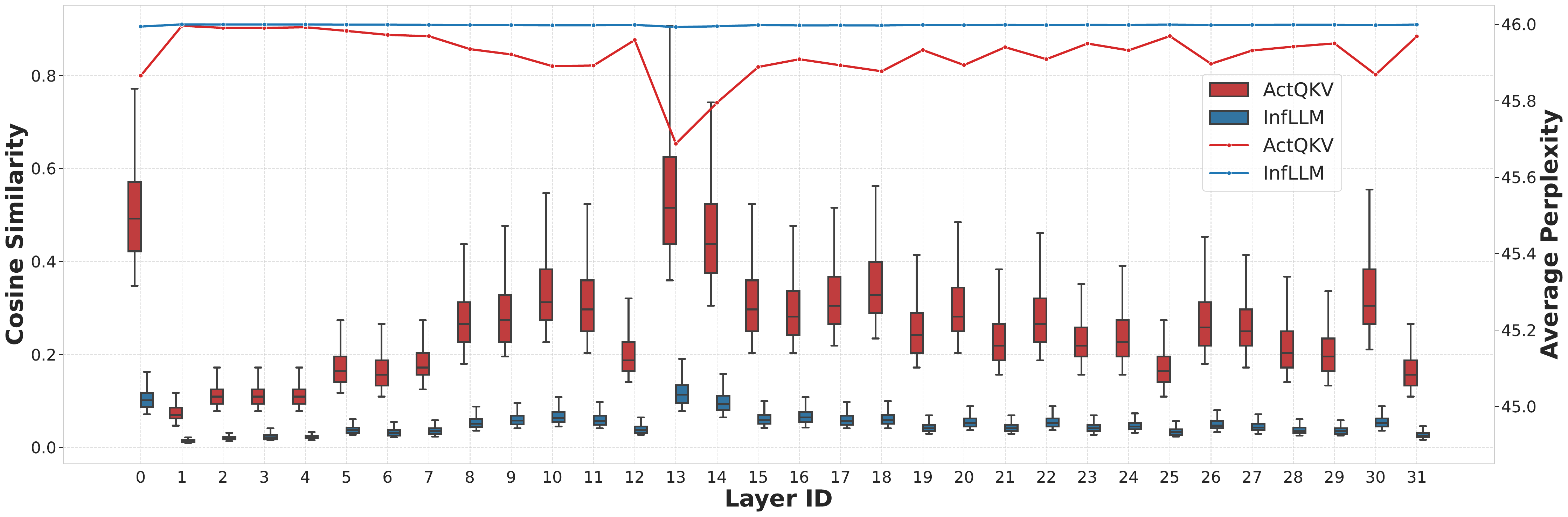}
    \caption{Analysis of the top-$k$  (avg. k=1,472) most relevant KV pairs for each inference step across layers. We randomly select 50 samples from Long-Bench and filter out those with a length less than 8K. In each layer, we calculate 35,180 similarity scores generated by our \name and InfLLM respectively. Each score is calculated based on a \pq and a chunk containing 32 KV pairs. The average perplexity is calculated based on the perplexity within the scores of each sample.
    } 
    \label{fig:analysis}
\end{figure*}

\subsection{Ablation Studies}
\label{sec:ablation}
In this subsection, we present ablation studies shown in \cref{tab:ablation} to evaluate two key components of our method: the Activation-aware Probe-Query $Q_\text{probe}^t$ (APQ, see \cref{sec:pq}) and the Dynamic Cut-off Mechanism (DCM, see \cref{sec:co}). 

When using APQ for key-value (KV) pair matching, our method attains a comparable score of 48.8, especially getting the best result 98.0 in retrieval tasks. These results demonstrate that the APQ component effectively captures the semantic context of the window for KV matching, outperforming conventional mean pooling approaches. Moreover, the incorporation of DCM, which dynamically determines the number of KV pairs to recall at each layer, further enhances the model’s ability of irrelevant information filtering. Overall, our approach employs a two-stage KV retrieval process following the traditional information retrieval paradigms: first, an initial retrieval stage identifies potentially relevant KV pairs; subsequently, a refined recall stage optimizes the selection process, achieving a peak performance of 49.4.

\subsection{Analysis of Retrieved KV Pairs}
\label{sec:analysis}

In this subsection, we compare the retrieved KV pairs from our \name and InfLLM methods to evaluate the specific impact of our proposed approach. To facilitate this comparison, we present the distribution of cosine similarity scores and average perplexity in \cref{fig:analysis} and analyze the following:

\paragraph{Cosine Similarity.} The box of cosine similarity clearly shows that ActQKV consistently achieves higher similarity scores across most layers compared to InfLLM. This outcome can be attributed to the activation-aware query (\pq) we introduced, which more effectively captures the underlying semantic information of the window context for each inference step. Furthermore, the enlargement of the box plots indicates that the distribution of similarities becomes more dispersed. This suggests that our \pq covers a broader semantic space, thereby resulting in a more robust KV retrieval process. The greater spread in the similarity values also reflects the model’s ability to account for a wider range of relevant KV pairs, ultimately enhancing the precision and adaptability of the retrieval process across different contexts.

\paragraph{Average Perplexity.} With respect to average perplexity, \name consistently shows lower perplexity scores compared to InfLLM which maintains a value of around 46.0. This indicates that \name yields more coherent and predictable results across the all layers. Notably, in layers 0 and 13, we notice significant differences, with \name showing more variation than InfLLM. This suggests that our retrieval method can flexibly adapt to the characteristics of different layers. By reducing perplexity, \name improves the ability to discriminate relevant KV pairs from irrelevant ones, resulting in more coherent and less uncertain historical information for long-context inferences in LLMs.

%% file: latex/tables/long.tex
\begin{table*}[!ht]
  \centering

\setlength{\tabcolsep}{1.8mm} 
\renewcommand{\arraystretch}{1} 
\scalebox{0.9}{
\begin{tabular}{l|c|cccccc>{\columncolor{teal!15}}c}
\toprule
Method & \basellama & Infinite & Stream & InfLLM & QLLM & TSLLM & EMLLM & \name \\
KV Budget & \multicolumn{1}{c|}{full context} & 2K & 2K & 2K & 2K & 2.5K & 4K & 2K \\
\midrule
NarrativeQA & 19.85 & 16.47 & 15.12 & 19.41 & 25.60 & 22.44 & 22.50 & \textbf{27.04} \\
Qasper &  42.36 & 32.01 & 31.72 & 41.27 & 39.12 & 40.74 & \textbf{44.95} & 40.42 \\
MultiFieldQA & 41.03 & 31.63 & 30.99 & 45.89 & 48.30 & 47.73 & 48.79 & \textbf{50.70} \\
HotpotQA &  47.38 & 34.73 & 35.26 & 44.97 & 49.91 & 50.33 & 49.19 & \textbf{51.37} \\
2WikiMQA & 39.20 & 29.22 & 30.59 & 36.27 & 39.63 & 31.38 & 38.08 & \textbf{42.07} \\
Musique & 22.96 & 13.50 & 13.64 & 19.73 & 25.03 & 24.53 & 25.19 & \textbf{33.40} \\
GovReport &  29.94 & 27.84 & 27.83 & 30.68 & 29.80 & \textbf{32.56} & 30.85 & 32.00 \\
QMSum &  21.45 & 19.91 & 20.14 & 21.36 & 22.23 & \textbf{23.50} & 22.77 & 23.06 \\
MultiNews & 27.51 & 27.36 & 27.37 & 27.87 & 27.85 & \textbf{27.92} & 27.28 & 27.26 \\
Trec & 74.00 & - & - & 57.50 & 55.50 & 67.50 & \textbf{73.50} & 69.50 \\
TriviaQA &  90.50 & \textbf{88.07} & 87.35 & 88.03 & 87.70 & \textbf{92.22} & 90.91 & 85.68 \\
SAMSum &  42.30 & 36.93 & 35.97 & 34.86 & 34.97 & 42.16 & \textbf{43.24} & 40.10 \\
PassageRetrieval & 62.50 & 23.50 & 23.50 & 85.25 & 88.00 & 87.00 & 86.00 & \textbf{94.50} \\
LCC &  60.83 & 60.42 & 58.15 & 58.17 & 58.37 & 58.86 & 60.44 & \textbf{62.04} \\
RepoBench-P & 49.14 & \textbf{64.95} & 62.97 & 62.01 & 61.04 & 51.24 & 44.88 & 61.92 \\
\midrule
Average & 44.73 & 36.18 & 35.76 & 43.98 & 46.20 & 46.67 & 47.24 & \textbf{49.40} \\
\bottomrule
\end{tabular}}
\label{tab:llama3-long}
\caption{
    \label{tab:llama3} Long-Bench (avg. 31K tokens)~\cite{longbench}.
     The comparison of results based on \basellama~\cite{llama3} are conducted from the works \cite{qllm, tokenselect, emllm}. Our results are highlighted in teal and best results are indicated in bold.
}
\vspace{-0.5em}
\end{table*}

%% file: latex/tables/infty.tex
\begin{table}[!ht]
  \centering
  \setlength{\tabcolsep}{1.5pt}
  \scalebox{0.8}{%
    \begin{tabular}{l|c|cccccc|c}
      \toprule
      \textbf{Method} & \textbf{KV} & \multicolumn{7}{c}{\textbf{$\infty$-Bench} (214K tokens)} \\
      \cline{3-9}
      & \textbf{Budget} & C.D. & M.F. & MC & R.KV & R.P & R.N & \textbf{Avg.} \\
      \midrule
      InfLLM  & 2k   & 22.59 & 26.86 & 33.19 & 80.80 & 100.0 & 28.64 & 48.68 \\
      QLLM    & 2k   & 23.10 & 27.37 & 34.50 & \textbf{84.00} & 100.0 & 27.63 & 49.43 \\
      TSLLM   & 2.5k & 27.41 & 28.29 & \textbf{45.85} & 40.00 & 100.0 & \textbf{97.29} & 56.47 \\
      EMLLM   & 8k   & 31.73 & 17.14 & 40.61 & 5.00  & 100.0 & 99.49 & 49.00 \\
      \rowcolor{teal!15} \name & 2k & \textbf{42.86} & \textbf{29.43} & 38.22 & 46.20 & \textbf{100.0} & 93.90 & \textbf{58.43} \\
      \bottomrule
    \end{tabular}%
  }
  \caption{\label{tab:llama3-infinitybench} $\infty$-Bench (avg. 122K tokens)~\cite{infinitebench}. The results comparison based on \basellama~\cite{llama3}. Our results are highlighted in teal and best results are indicated in bold.}
\end{table}

%% file: latex/tables/ablation_study.tex
\begin{table}[!t]
  \centering
  \setlength{\tabcolsep}{2.5pt}
  \scalebox{0.8}{
    \begin{tabular}{l|c|cccccc|c}
      \toprule
      {\textbf{Method}} & {\textbf{KV}} & \multicolumn{7}{c}{\textbf{LongBench}} \\
      \cline{3-9}
       & \textbf{Budget} & SQA & MQA & Sum & FSL & Ret & Cod & \textbf{Avg.} \\
      \midrule
      TSLLM & 2.5k & 37.0 & 35.4 & 28.3 & \textbf{67.3} & 87.0 & 51.2 & 46.7 \\
      EMLLM & 8k & 39.3 & 37.7 & 27.0 & \textbf{69.2} & 87.5 & 50.3 & 47.2 \\
      w/ APQ & 2k & \textbf{40.3} & 40.7 & \textbf{27.5} & 63.1 & \textbf{98.0} & 61.5 & 48.8 \\
      w/ DCM & 2k & 39.7 & 42.1 & 27.4 & 64.3 & 94.5 & 61.7 & 49.2 \\
      \name & 2k & 39.4 & \textbf{42.3} & 27.4 & 65.1 & 94.5 & \textbf{62.0} & \textbf{49.4} \\
      \bottomrule
    \end{tabular}
}
\caption{
    \label{tab:ablation}
    The ablation study of our method \name, where \textbf{A}ctivated \textbf{P}robe-\textbf{Q}uery for KV matching and \textbf{D}ynamic \textbf{C}ut-off \textbf{M}echanism  for KV recall. We use the mean pooling to represent \pq in w/ APQ as same as InfLLM and QLLM.
}
\end{table}

%% file: latex/tex/conclusion.tex
\section{Conclusion}

In this paper,  we present \name, a training-free method to KV retrieval efficiency for long-context LLMs inference. The primary challenge in KV retrieval stems from the inherent vagueness  of existing \pq, which inadequately filter irrelevant KV pairs. To address this limitation, we develop an activation-aware \pq construction strategy and a layer-wise KV cut-off mechanism to effectively match and recall  the relevant KV pairs. We hope this work can inspire the broader research for  LLMs representation methods, leading to improved long-context information filtering capabilities akin to specialized embedding models.

%% file: latex/tex/limitations.tex
\section*{Limitations}
Our method achieves promising performance to enhance the relevant KV pairs retrieval for long-context LLMs inference. And we believe that the interpretability of the retrieved KV pairs requires further exploration in future works. Unlike non-autoregressive architectures in embedding models, the auto-regressive architecture of LLMs results in the semantics of current tokens being influenced by historical KV pairs. When processing a long context all at once, this interaction makes it difficult to separate the semantics from various events because the retrieved key-value pairs mostly show historical information. This introduces challenges in interpreting the retrieval results.

%% file: latex/tex/appendix.tex
\appendix
\label{sec:appendix}
\input{latex/tables/complexity}
\begin{table*}[!t]
\centering  
\resizebox{0.95\textwidth}{!}{
\begin{tabular}{lclrccc}
\toprule
Dataset & ID & Source & Avg len & Metric & Language & \#data \\
\midrule
\emph{Single-Document QA} \\
NarrativeQA & 1-1 & Literature, Film & 18,409 & F1 & English & 200 \\
Qasper & 1-2 & Science & 3,619 & F1 & English & 200 \\
MultiFieldQA-en & 1-3 & Multi-field & 4,559 & F1 & English & 150 \\

\midrule
\emph{Multi-Document QA} \\
HotpotQA & 2-1 & Wikipedia & 9,151 & F1 & English & 200 \\
2WikiMultihopQA & 2-2 & Wikipedia & 4,887 & F1 & English & 200 \\
MuSiQue & 2-3 & Wikipedia & 11,214 & F1 & English & 200 \\

\midrule
\emph{Summarization} \\
GovReport & 3-1 & Government report & 8,734 & Rouge-L & English & 200 \\
QMSum & 3-2 & Meeting & 10,614 & Rouge-L & English & 200 \\
MultiNews & 3-3 & News & 2,113 & Rouge-L & English & 200 \\

\midrule
\emph{Few-shot Learning} \\
TREC & 4-1 & Web question & 5,177 & Accuracy (CLS) & English & 200 \\
TriviaQA & 4-2 & Wikipedia, Web & 8,209 & F1 & English & 200 \\
SAMSum & 4-3 & Dialogue & 6,258 & Rouge-L & English & 200 \\

\midrule
\emph{Retrieval} \\

PassageRetrieval-en & 5-1 & Wikipedia & 9,289 & Accuracy (EM) & English & 200 \\

\midrule
\emph{Code Completion} \\
LCC & 6-1 & Github & 1,235 & Edit Sim & Python/C\#/Java & 500 \\
RepoBench-P & 6-2 & Github repository & 4,206 & Edit Sim & Python/Java & 500 \\
\bottomrule
\end{tabular}
}
\caption{An overview of the dataset statistics in LongBench~\cite{longbench}. `Avg len' (average length) is computed using the number of words for the English (code) datasets and the number of characters for the Chinese datasets. `Accuracy (CLS)' refers to classification accuracy, while `Accuracy (EM)' refers to exact match accuracy.}
\label{tab:dataset-long}
\vspace{-5mm}
\end{table*}

\newpage

\section{The Complexity of LLMs Inference}

In this section, we focus on the attention computation and analyze the complexity of exiting methods shown in \cref{tab:complexity} as follows:

\textbf{Standard Attention Mechanism.}
Under the standard attention mechanism, during the pre-filling stage, each token in the input sequence undergoes attention calculations with all other tokens, resulting in a time complexity of $O(N^2)$. In the decoding stage, as the context grows, the complexity of generating each new token increases accordingly. When generating the $t$-th token, the length of the context to be processed is $N+t$, so the total time complexity of the decoding stage is $O(\sum^{M}_{t=1}M(N+t)^2)$, which is approximately $O(N^2M+M^3)$. Since the decoding length $M$ is usually much smaller than the input sequence length $N$, the overall complexity can be simplified to $O(N^2+MN^2)$.

\textbf{Sliding Window Mechanism with KV Cache.}
The sliding window mechanism divides the input sequence into several windows of fixed size, each with a size of $m$. During the pre-filling stage, the processing complexity of the tokens within each window is $O(m^2)$, and the interaction complexity between the KV caches of the windows is approximately $O(N)$, so the overall time complexity is $O(\frac{N}{m} \times m^2) = O(mN)$, which is equivalent to $O(N^2)$ when the window size m is constant and linearly dependent on N. In the decoding stage, the decoding of each new token only needs to interact with the m tokens in the current window and some tokens in the adjacent windows, resulting in a total time complexity of $O(MN)$. Overall, the time complexity can be simplified to $O(N^2+MN)$.

\begin{figure*}[!ht] 
    \centering
    \begin{minipage}{0.45\textwidth}
        \centering
        \includegraphics[width=\textwidth]{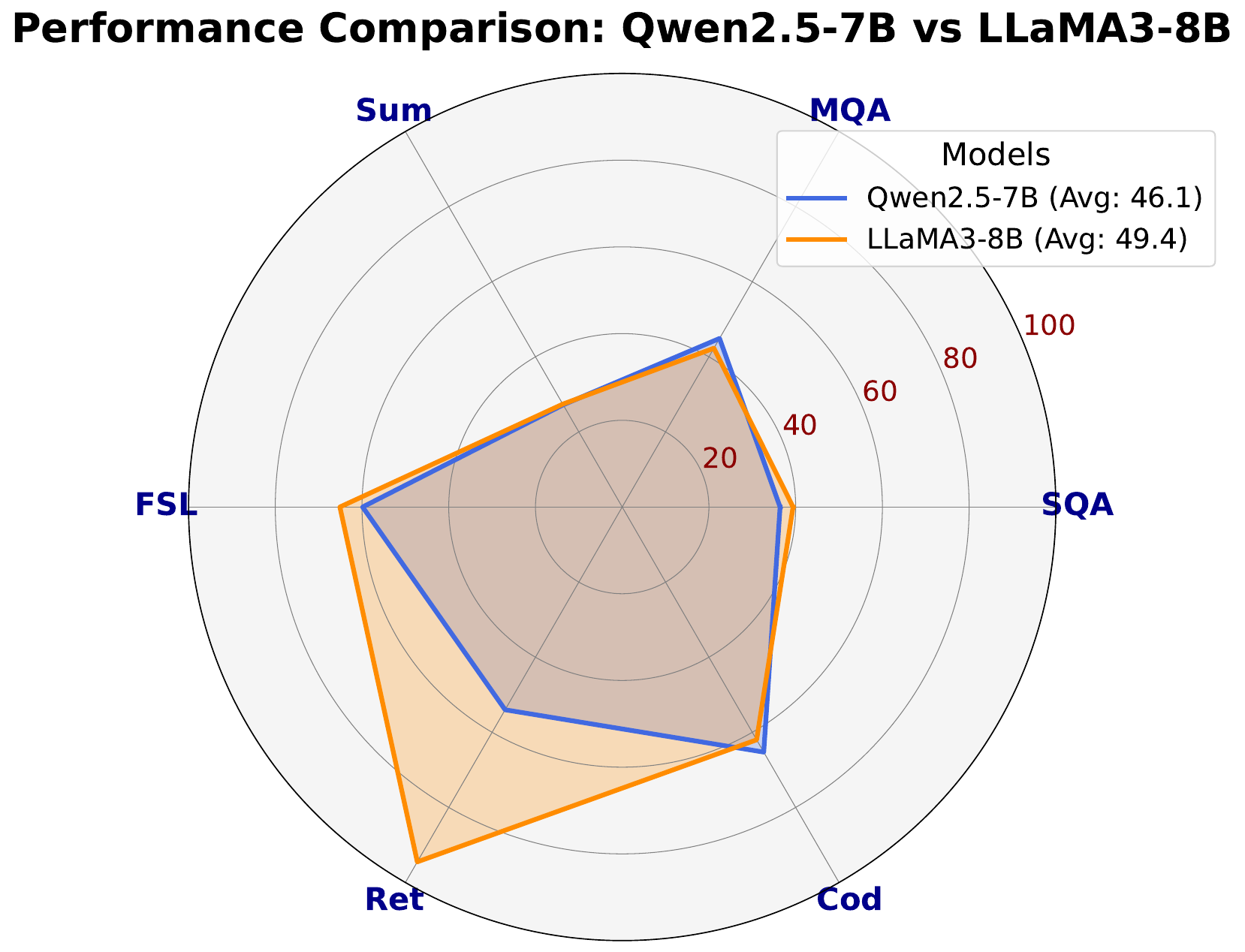}
        \subcaption{Long-Bench~\cite{longbench}.}
        \label{fig:radar_long}
    \end{minipage}
    \hfill
    \begin{minipage}{0.45\textwidth}
        \centering
        \includegraphics[width=\textwidth]{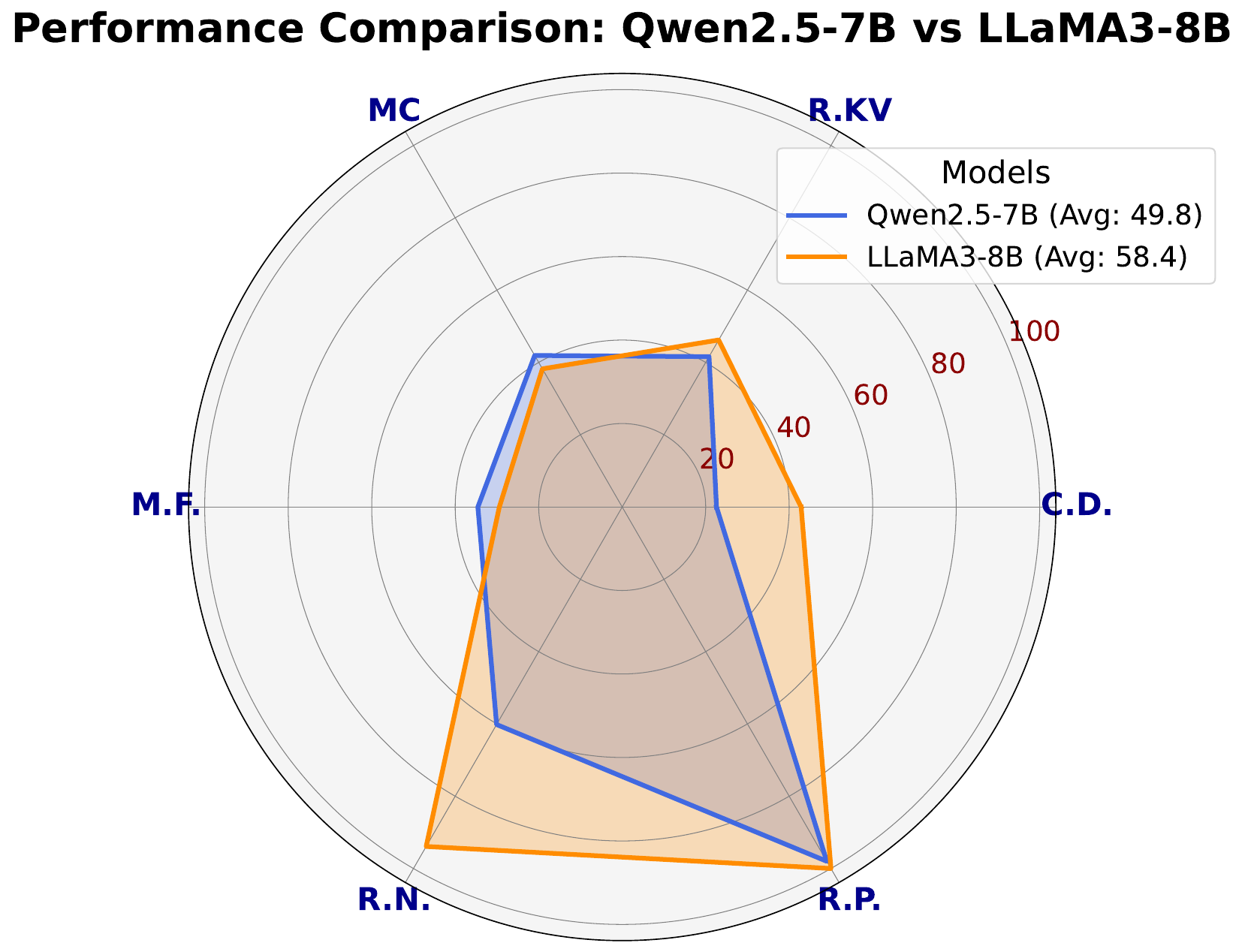}
        \subcaption{$\infty$-Bench~\cite{infinitebench}.}
        \label{fig:radar_infty}
    \end{minipage}
    \caption{Performance comparison based on different models: LLaMA3-8B~\cite{llama3} v.s. Qwen2.5-7B~\cite{qwen2.5}.}
    \label{fig:models}
\end{figure*}
\textbf{KV Retrieval for Long-Context Inference.}
When using the method of Top-k retrieval combined with the sliding window, the pre-filling stage divides the input sequence into windows of fixed size. During the processing of the tokens in each window, only the interaction with the top k most relevant key-value pairs is performed, so the complexity of the pre-filling stage is $O(\frac{N}{m}\times(m+k))$, which can be approximated as $O(kN)$ if the window size m and the retrieval range k meet certain conditions. In the decoding stage, the prediction of each new token only needs to interact with the top-$k$ most relevant key-value pairs, with a time complexity of $O(kM)$. Overall, the time complexity is simplified to $O(kN+kM)$.

\begin{table}[!ht]
    \footnotesize
    \centering
    \begin{tabular}{l|cccc}
        \toprule
        \textbf{Task}  & \textbf{Annotation}  & \textbf{\# Ex.} & \textbf{Avg Len} \\
        \midrule
        Ret.PassKey    & Auto & 590 & 122.4K/2\\
        Ret.Number     & Auto & 590 & 122.4K/4\\
        Ret.KV         & Auto & 500 & 121.1K/22.7\\
   
        En.MC       & Human & 229 & 184.4K/5.3\\

        Code.Debug  & Human & 394 & 114.7K/4.8\\

        Math.Find   & Auto & 350 & 87.9K/1.3 \\
        \bottomrule
    \end{tabular}
    \caption{Data statistics of $\infty$-Bench~\cite{infinitebench}. The columns indicate whether the annotation was auto-generated or done by humans, the number of examples, and the average length (input/output) in tokens.}
    \label{tab:dataset-infty}
\end{table}

\begin{figure*}[!ht] 
    \centering
        \includegraphics[width=1.0\textwidth]{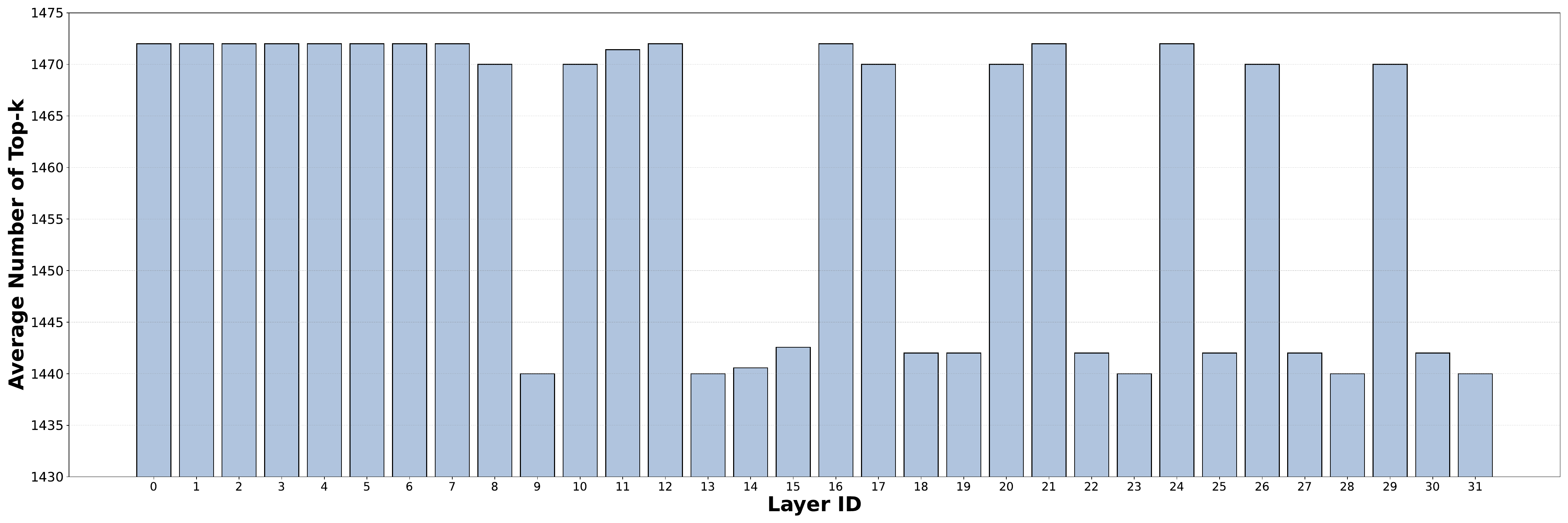}
    \caption{Average number of relevant KV pairs recalled for each  layer in decoding stage based on \basellama~\cite{llama3}. We
randomly select 50 samples from Long-Bench and filter out those with a length less than 8K. } 
    \label{fig:recall}
\end{figure*}
\section{Details in Long-Bench and $\infty$-Bench}
\label{sec:alg}
Long-Bench (95\% sequence length is 32K) focuses on tasks that involve reasoning, such as question answering, summarization, few-shot learning, retrieval, and coding.  The groups of datasets are categorized as follows: \textbf{Single-doc QA}: NarrativeQA, Qasper, MultiFieldQA; \textbf{Multi-doc QA}: HotpotQA, 2WikiMQA, Musique; \textbf{Summarization}: GovReport, QMSum, MultiNews; \textbf{Few-shot Learning}: TREC, TriviaQA, SAMSum; \textbf{Retrieval}: PassageRetrieval; \textbf{Code}: RepoBench-P. And $\infty$-Bench (avg. length of 200K) emphasizes factual retrieval, covering domains such as code, mathematics, multiple-choice questions, and general retrieval tasks. The statistics and evaluation metrics of datasets are detailed in ~\cref{tab:dataset-long} and ~\cref{tab:dataset-infty}.

\section{Experimental Results}
\label{sec:experiments}
All experiments were implemented using PyTorch and performed on two NVIDIA A800 80GB GPUs. In all experiments in this paper, we use standard greedy decoding to ensure reliable results.

\subsection{Model Comparison}
We conduct experiments on Long-Bench~\cite{longbench} and $\infty$-Bench~\cite{infinitebench} using LLaMA3-8B~\cite{llama3} and Qwen2.5-7B~\cite{qwen2.5}, as illustrated in ~\cref{fig:models}. 

For LLaMA3-8B~\cite{llama3}, the model achieves state-of-the-art (SOTA) performance across tasks in both Long-Bench and $\infty$-Bench, demonstrating its versatility, particularly in factual retrieval and code-related tasks.
In contrast, although Qwen2.5-7B~\cite{qwen2.5} does not match the performance of LLaMA3-8B across all categories, it exhibits substantial improvements over the baseline. The most significant performance drop is observed in the Retrieval tasks, where Qwen2.5-7B underperforms relative to LLaMA3-8B. This highlights a challenge in handling retrieval-related aspects of the benchmarks. Nevertheless, Qwen2.5-7B consistently outperforms the baseline in these tasks, underscoring the effectiveness of our approach, even though it does not yet match the top-performing model in retrieval. However, Qwen2.5-7B excels in code-related tasks, even surpassing LLaMA3-8B in this domain. This demonstrates the model’s proficiency in handling complex, domain-specific tasks, such as those encountered in RepoBench-P. While Qwen2.5-7B shows some weaknesses in retrieval, its performance in other specialized areas is either competitive or superior.

In total, although Qwen2.5-7B experiences a decline in retrieval task performance compared to LLaMA3-8B, it still outperforms the baseline, validating the effectiveness of our method, \name.

\subsection{Dynamic KV Pairs Recall}

Our approach employs a layer-wise key-value cut-off mechanism and an activation-aware \pq construction strategy to more effectively match and recall relevant KV pairs. As shown in ~\cref{fig:recall}, we report the average number of relevant KV pairs recalled for each layer.

The results in ~\cref{fig:analysis} demonstrate that our method, \name, adapts to the varying distributions across layers, ensuring a robust and efficient retrieval process. Notably, in layer 13, which exhibits the lowest perplexity of similarity scores and receives the smallest KV budget, our method fully aligns with the objectives outlined in ~\cref{eq:budget}. This consistency allows LLMs to effectively process long-context information for long-context inference.

%% file: latex/tables/complexity.tex
\begin{table*}[!ht]
\centering
\setlength{\tabcolsep}{1.8mm} 
\renewcommand{\arraystretch}{1} 
\scalebox{0.9}{
\begin{tabular}{l|c|c|c|}
\hline
\textbf{Mechanism} & \textbf{Pre-filling Complexity} & \textbf{Decoding Complexity} & \textbf{Overall Complexity} \\
\hline
Standard Attention & $O(N^2)$ & $O(N^2 + MN^2)$ & $O(N^2 + MN^2)$ \\
\hline
Sliding Window with KV Cache & $O(N^2)$  & $O(N^2 + MN)$ & $O(N^2 + MN)$ \\
\hline
\rowcolor{teal!15}  KV Retrieval for Long-Context Inference & $O(kN)$ & $O(kN + kM)$ & $O(kN + kM)$ \\
\hline
\end{tabular}}
\caption{Complexity analysis of different methods. Our \name is belong to the KV retrieval method and the complexities are highlighted in teal.}
\label{tab:complexity}
\end{table*}